\documentclass[10pt,twocolumn,letterpaper]{article}

\usepackage{iccv}
\usepackage{times}
\usepackage{epsfig}
\usepackage{graphicx}
\usepackage{amsmath}
\usepackage{amssymb}
\usepackage{mathtools}
\usepackage{commath}
\usepackage{subfigure}
\usepackage{textcomp}
\usepackage{booktabs} 
\usepackage{float} 
\usepackage{bm}
\usepackage{algorithm}
\usepackage{algorithmic}
\usepackage[pagebackref=true,breaklinks=true,colorlinks,bookmarks=false]{hyperref}

\iccvfinalcopy

\def\iccvPaperID{9615}

\ificcvfinal\fi
\begin{document}

\title{Online Preconditioning of Experimental Inkjet Hardware\\ by Bayesian Optimization in Loop}

\author{Alexander E. Siemenn\\
MIT\\
Department of Mechanical Engineering\\

{\tt\small asiemenn@mit.edu}

\and
Matthew Beveridge\\
MIT\\
Department of EECS\\
{\tt\small mattbev@mit.edu}\\

\and
Tonio Buonassisi\\
MIT\\
Department of Mechanical Engineering\\
{\tt\small buonassisi@mit.edu}\\

\and
Iddo Drori\\
MIT\\
Department of EECS\\
{\tt\small idrori@mit.edu}
}

\maketitle

\begin{abstract}

High-performance semiconductor optoelectronics such as perovskites have high-dimensional and vast composition spaces that govern the performance properties of the material. To cost-effectively search these composition spaces, we utilize a high-throughput experimentation method of rapidly printing discrete droplets via inkjet deposition, in which each droplet is comprised of a unique permutation of semiconductor materials. However, inkjet printer systems are not optimized to run high-throughput experimentation on semiconductor materials. Thus, in this work, we develop a computer vision-driven Bayesian optimization framework for optimizing the deposited droplet structures from an inkjet printer such that it is tuned to perform high-throughput experimentation on semiconductor materials. The goal of this framework is to tune to the hardware conditions of the inkjet printer in the shortest amount of time using the fewest number of droplet samples such that we minimize the time and resources spent on setting the system up for material discovery applications. We demonstrate convergence on optimum inkjet hardware conditions in 10 minutes using Bayesian optimization of computer vision-scored droplet structures. We compare our Bayesian optimization results with stochastic gradient descent.

\end{abstract}

\section{Introduction}
\begin{figure*}[t]
\centering
\includegraphics[width=.75\textwidth]{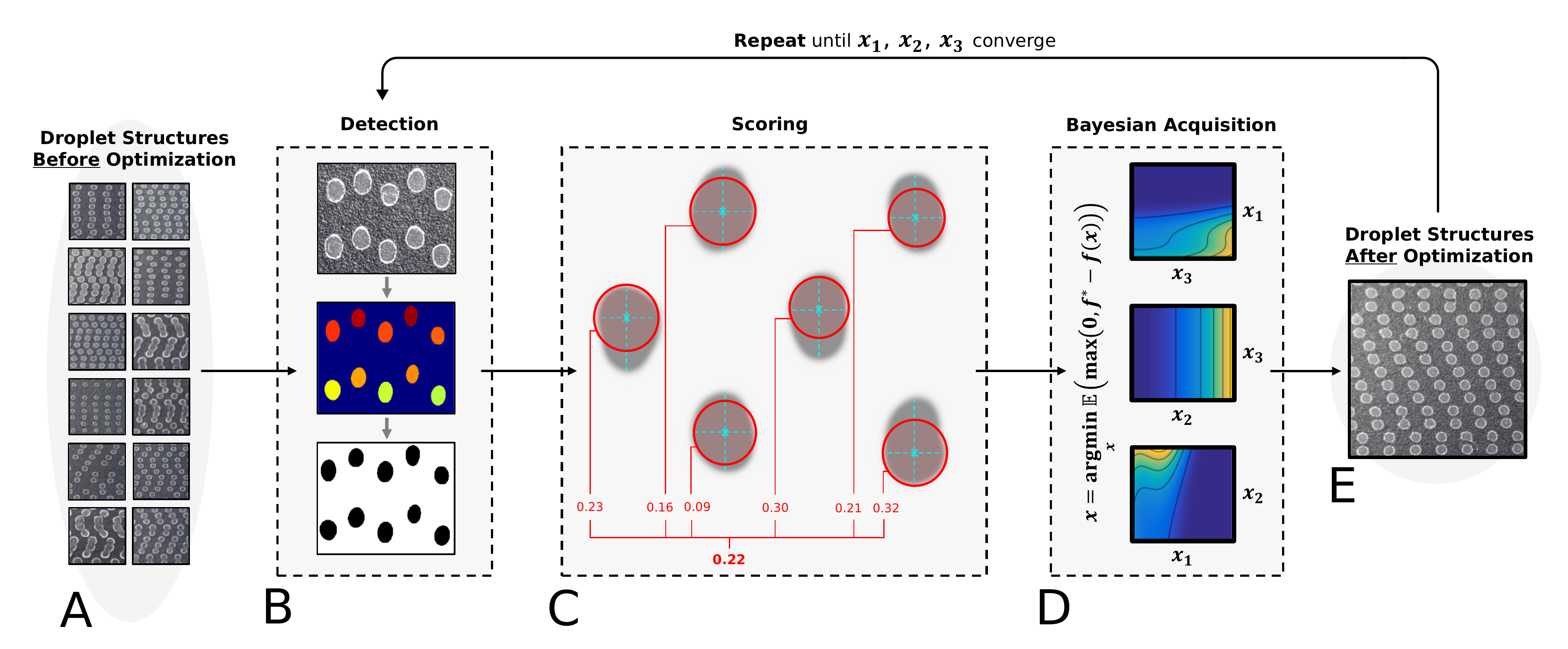}
\caption{Process flow for preconditioning inkjet hardware to reliably generate droplet structures of high geometric uniformity and high yield. (A) An initialization dataset of 12 droplet structures is synthesized using hardware printing conditions obtained from LHS. (B) Images of droplet structures are segmented from the background substrate using a watershed method. The raw image (top) is segmented into indexed droplets (middle) which is then converted into a binary segmentation of droplet and background pixels (bottom) (C) The detected features are assigned a loss score to each droplet structure by computing yield losses and geometric losses. Yield losses are a function of the ratio of droplet pixels to background pixels and number of segmented droplets in the sample. Geometric losses are shown in (C) where each droplet is assigned a loss depending on how far the droplet shape deviates from a perfect circle mapped to its centroid. (D) The loss score obtained is used to label each hardware printing condition such that BO suggests new printing conditions that minimize the loss score. The expected improvement acquisition function is used to search the parameter space for these conditions. (E) Hardware printing conditions suggested by BO are used to synthesize a new droplet structure sample. This sample is fed back into the model and the process is repeated online in a loop until we converge on hardware printing conditions that reliably generate optimized droplet structures.}
\label{fig:workflow}
\vspace{-10pt}
\end{figure*}

There are hundreds of thousands of chemical compositions that may form high-performance optoelectronics such as perovskites \cite{Sun2020}. It is not experimentally tractable to search this vast composition space for compositions with optimized performance properties (\textit{e.g.} high efficiency and low degradation) by synthesizing full perovskite cells via conventional sputtering or spray coating methods due to the time and resources consumed by this process \cite{Sun2019,Ren2020, Boyd2019,Sampaio2020,Zhuk2017}. To synthesize a perovskite solar cell via conventional methods takes approximately 1 hour per cell and consumes roughly 40mg of costly precursor material. Hence, searching this vast composition space via conventional synthesis methods is highly time and cost inefficient. Therefore, we look towards a method of efficient high-throughput experimentation in which we reduce our samples to the most fundamental form -- droplets -- which are still characterizable to determine the performance properties of the composition \cite{Bash2020}. Utilizing conventional inkjet deposition methods elicits this high-throughput experimental exploration of semiconductor composition spaces. However, inkjet deposition methods are not inherently optimized to generate semiconductor droplets that can be reliably characterized. For droplets of semiconductor material to be reliably characterized they must have (1) high geometric uniformity and (2) high yield. Geometric uniformity ensures all droplets are characterized under the same conditions to reduce variability. Yield ensures each droplet has enough material to characterize while maximizing the droplet count per unit area. Setting up a printer to meet these two objectives requires domain knowledge of the printed fluid response to changes in hardware printing conditions as well as time and material resources to explore this domain to discover the optimized hardware printing conditions for semiconductor experimentation. Thus, to minimize the time and resources required to find hardware printing conditions that optimize printed droplet structures without the knowledge of a domain expert, we propose a method of computer vision-driven Bayesian optimization that iteratively updates the hardware conditions of a printer until we converge on conditions that reliably generate optimized droplet structures.

We utilize image segmentation and processing methods to detect the droplets in each sample printed by the inkjet printer. The inkjet printer synthesizes droplets by tuning three parameters: (1) jetting pressure, (2) valve actuation frequency, and (3) nozzle speed. A two-level model segments and then scores each droplet structure based on our defined objective function to (1) maximize droplet uniformity and (2) maximize droplet yield \cite{deneault2020, opencv_library}. Bayesian optimization (BO) is an effective supervised method of machine learning (ML) that is used to discover an optimum condition from a state space when the objective function is costly to solve \cite{Gelbart2014}. In this paper, BO is utilized to efficiently search for printer conditions that minimize the loss score of our droplet structures \cite{rasmussen2003gaussian, bergstra2012, snoek2012}. New optima are synthesized iteratively in a loop until we converge on printing conditions that reliably generate optimized droplet structures. The performance of BO in loop is benchmarked against a stochastic gradient descent (SGD) model, which is the backbone of many conventional ML algorithms \cite{Kiwiel2001}.

The objective of this work is to develop a process optimization method for preconditioning inkjet printer hardware such that high-throughput experimentation of high-performance semiconductor compositions, such as perovskites, is enabled without the intervention of a domain expert. By utilizing ML in loop with experimental synthesis, time and reagent resources are saved since optimum convergence on hardware printer conditions is attained using fewer synthesized samples relative to pure experimental exploration \cite{Sun2020, Ren2020, Sun2019}. Demonstrating automated, fast, and economical optimum convergence on hardware conditions elicits significant implications for the high-throughput engineering of stable, high-performance semiconductor optoelectronics \cite{Boyd2019}.

\subsection{Related Work}
Previous work has been conducted within the realm of combining ML with experimental synthesis to discover semiconductor material compositions, such as perovskites, that minimize degradation \cite{Ren2020a,Ren2020,Sun2020,Langner2020,Sun2019}. An accomplishment of the previous work of Sun \etal \cite{Sun2020} is in the tuning of the BO algorithm to accelerate the convergence of a stable perovskite composition in three iterations using only 112 experimental samples in a search space of $5\times10^5$ with only 15 initial training samples. The work of Sun \etal \cite{Sun2020, Sun2019} and Ren \etal \cite{Ren2020} demonstrate the application of this ML and experimentation fusion tool on perovskites to quickly and cost-effectively discover optimum material compositions that optimize a desired material property. Our paper builds off of these methods whereby instead of aiming to discover an optimum material composition that optimizes a material property, we use this ML and experimentation fusion tool to discover optimum device hardware settings that prepare our hardware for experimental synthesis.

ML in loop has proven to be a useful catalyst in other application settings as well; notably, recent work demonstrates holographic storage as a viable improvement to traditional spinning disk drives and solid-state drives \cite{Bishop2020, chatzieleftheriou2020could}. Using interfering beams of light, Chatzieleftheriou \etal \cite{chatzieleftheriou2020could} change the electronic structure of a crystal and store overlapping pieces of data within it, thus, creating a  write-once-read-many, vision-based storage device. This is possible by using a convolutional neural network (CNN) in loop to minimize error when reading and writing user data. Iteratively using a CNN in this case converges to the optimal conditions for embedding holograms quicker than when using more traditional synthesis methods. A similar approach proves effective for inkjet printing of droplet structures.

An experimental and computational fusion tool has also been demonstrated to reduce the number of material samples necessary to discover the optimum synthesis conditions for additively manufactured structures \cite{Gongora2020}. By measuring the force-displacement relationship of only 240 additively manufactured structures, Gongora \etal \cite{Gongora2020} shows convergence to an optimized structure that maximizes toughness. Without using the computational power of BO in tandem with experimental synthesis, it would require 60x more experiments to converge to this optimized structure. From this related work, data fusion BO demonstrates the potential to reduce the of experiments necessary to discover optimum hardware settings to precondition inkjet experimental setups for high-throughput semiconductor research.

\section{Methods}
Optimized inkjet hardware conditions are obtained by optimizing droplet structures through a computer vision-driven ML in loop workflow using only 12 training samples. The inkjet hardware optimized in this work and used to experimentally synthesize the droplet structures is shown in Figure \ref{fig:inkjet}. Our workflow is summarized in pseudo-code in Algorithm \ref{alg:experimental-process} and illustrated in Figure \ref{fig:workflow}.

\begin{algorithm}[t]
\small
   \caption{Droplet Optimization by SGD \& BO in Loop}
   \label{alg:experimental-process}
\begin{algorithmic}
   \STATE {\bfseries Input:} Initialized 3-dimensional printing condition parameter space via Latin Hypercube Sampling (LHS).
   \STATE {\bfseries Output:} Optimized parameter space.
   \STATE Define $S$, the set of experimentally synthesized training samples, initialized with printed samples from the input LHS parameter space.
   \STATE Let $\ell(\cdot) = \frac{w_g \mathcal{L}_{g_i} + w_e \mathcal{L}_{e_i}}{w_g+w_e}$, scoring function from Section \ref{sec:scoring-function}.
   \STATE Let $L = \ell(s) \; \forall s \in S$, the loss scores of the training samples.
  \WHILE{no convergence}
       \STATE \textbf{Train:} Train SGD and BO in parallel with $S$ and $L$.
       \STATE \textbf{Predict:} Using each SGD and BO predict the optimum printing conditions, $y_{\textrm{pred}}$, which minimize $\ell$.
       \STATE \textbf{Synthesize:} Print a sample $s_{\textrm{pred}}$ using conditions $y_{\textrm{pred}}$.
       \STATE \textbf{Update:} Add $s_{\textrm{pred}}$ to $S$, and recompute $L$. 
  \ENDWHILE
   \STATE \textbf{return} $y_*$.
\end{algorithmic}
\end{algorithm}

\subsection{Inkjet Printer Hardware}
\label{sec:hard_ink}
\begin{figure}
\centering
\subfigure[Inkjet Printer Hardware]{
\includegraphics[width=0.45\textwidth]{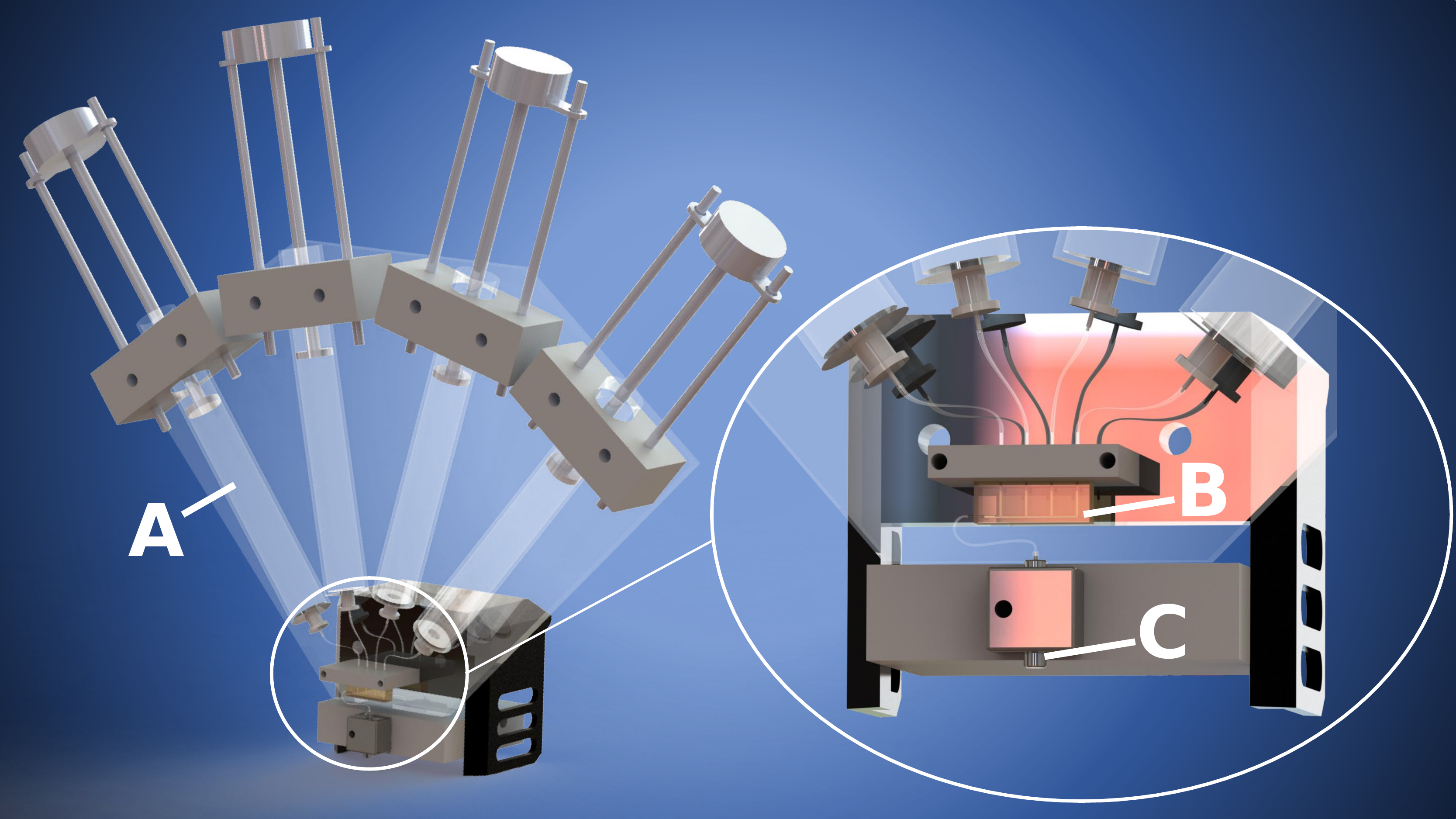}}
\label{fig:inkjet}
\subfigure[Droplet Structure Training Data]{
\includegraphics[width=0.45\textwidth]{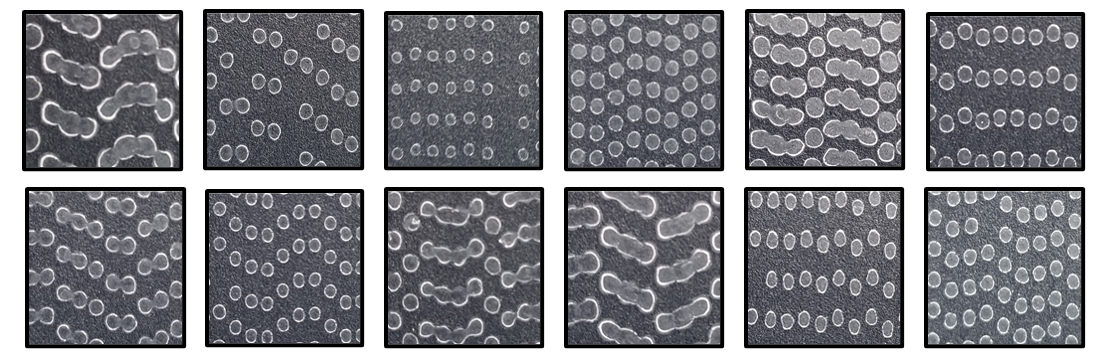}}
\label{fig:training_images}
\caption{(a) The developed inkjet system is shown here that is capable of synthesizing semiconductor compositions from up to four non-viscous precursor materials. The two core deviations of this hardware from traditional inkjet hardware are shown in (A) and (B). (A) The storage vials for the precursor material. (B) The precursor mixing point engineered to induce turbulent mixing. (C) An off-the-shelf piezoelectric-actuated inkjet nozzle where fluid is jetted once mixed. (b) By tuning the jetting pressure, valve actuation frequency, and nozzle translation speed printing conditions of the printer, different droplet structures are generated. The conditions used to synthesize the 12 samples shown here were obtained through LHS and are used to train the BO and SGD algorithms.}
\vspace{-10pt}
\end{figure}

To cost-effectively search the composition space of high-performance semiconductor-based optoelectronics, a printer system is fabricated using traditional inkjet hardware with modifications\footnote{A video of this hardware operating is found \href{https://www.dropbox.com/s/zsuf1x9x9lisufu/inkjet_droplet_print.mp4?dl=0}{here}} to enable printing of several compositions which is combined at a mixing point. For this study, we focus on the process optimzation preconditioning of the hardware to generate droplets optimized for semiconductor research. To precondition the hardware, dyed water is used to form the droplet structures such that the droplets are easily detectable by image processing. We developed our own inkjet hardware, which is shown in Figure \ref{fig:inkjet}; this hardware translates laterally to deposit droplet structures. Existing literature supports the use of the inkjet printing method to deposit functional semiconductor material onto a substrate, however, these studies do not address the prospect of using machine learning and computer vision to drive process optimization of hardware conditions necessary to precondition experimentation \cite{Calvert2001,Raut2018,Kassem2018,Glasser2019,Yun2009}. Using this hardware construction, three hardware printer parameters govern the structure of deposited droplets:

\textbf{Jetting pressure}. 
The driving printing parameter for how fast the material stream is ejected from the nozzle is the jetting pressure. By changing the pressure within the nozzle tubes, the material flow rate changes. Jetting pressure ranges between 0.02MPa-0.15MPa, and as the pressure increases, the amount of material ejected from the nozzle tip increases per unit time.

\textbf{Valve actuation frequency}.
The rate of droplet formation is driven by a valve actuation frequency parameter. The piezoelectric valve internal to the nozzle actuates at a frequency of 20Hz-40Hz which breaks the material stream into discrete droplets. As frequency increases, droplets are created at a faster rate.

\textbf{Nozzle translation speed}. 
The point of droplet deposition onto the substrate is governed by the nozzle translation speed. The nozzle translates parallel to the the deposition site in two-dimensions at speeds ranging between 300mm/s-900mm/s. As the nozzle translation speed increases, the droplets are deposited further apart and with more entropy.

\subsection{Detection and Scoring}
\label{sec:scoring-function}
A two-level model is used to (1) detect and (2) score the droplet structure samples experimentally synthesized by the inkjet hardware illustrated in Section \ref{sec:hard_ink}. To detect the droplet structures from each sample, $i$, a watershed image segmentation method is used to segment droplet pixels from background substrate pixels \cite{opencv_library}. To score the droplet structures from each sample, a linear combination of scalarized losses is computed. The first component of the score is the geometric loss -- quantified by the normalized sum of pixels that do not fit to a perfect circle mapped onto the centroid of each droplet \cite{deneault2020}:
\begin{equation} 
\label{geometric}
    \mathcal{L}_{g_i} = \sum \frac{\overline{P_{\textrm{droplet}} \cap P_{\textrm{circle}}}}{P_{\textrm{droplet}}},
\end{equation}

where $P_{\textrm{droplet}}$ are the pixels of a droplet and $P_{\textrm{circle}}$ are the pixels of a circle mapped to that droplet.

The second component of the score is yield loss -- quantified by the ratio of all non-droplet pixels to the total pixels:

\begin{equation} 
\label{connectivity}
    \mathcal{L}_{e_i} = \frac{\overline{\sum P_{\textrm{droplet}} }}{P_{\textrm{total}_i}},
\end{equation}
where $P_{\textrm{droplet}}$ are the pixels of a droplet and $P_{\textrm{total}}$ are the total number of pixels in the sample.

By combining these components, a loss score is defined for each sample image within the set: 
\begin{equation} 
\label{cv-loss}
    L_i=\frac{w_g \mathcal{L}_{g_i} + w_e \mathcal{L}_{e_i}}{w_g+w_e} ,
\end{equation}
where $\mathcal{L}_{g_i}, \mathcal{L}_{e_i} \in [0,1]$ are the geometric and yield loss scores for each image, respectively, and $w_g$ and $w_e$ are the geometric and yield-component weights, respectively. By tuning $w_g$ and $w_e$, the objective function adjusts to optimize for either a more geometric uniformity or higher droplet yield structure. In this study, we aim to minimize the value of $L_i$ using $w_g=w_e$ such that we attain both the most geometrically uniform and highest yield droplet structure. Obtaining a geometrically uniform and high-yield droplet structure is important for inkjet experimentation of semiconductors because it ensures that all droplets have enough material to be characterized, are characterized under the same conditions, and maximize the characterizable droplets per unit area. Computed droplet structures scores for the training samples used in this study are shown in Table \ref{samples}.

\subsection{Training}
Training our machine learning models begins by assigning each sample with a score using Equation \ref{cv-loss} that indicates how closely the sample corresponds to a geometrically uniform and high-yield structure. Both the BO and SGD models operate using the same initialization data set of 12 samples, which is obtained through Latin Hypercube Sampling (LHS) of the pressure $\times$ frequency $\times$ translation speed parameter space. LHS attempts to capture the most variability of an $N$-dimensional manifold using the fewest samples relative to other conventional sampling methods such as Monte Carlo \cite{mckay1979}. The training data samples are shown in Figure \ref{fig:training_images} with their corresponding values in Table \ref{samples}.

\begin{table}
\small
\caption{The 12 LHS-initialized sample printing conditions with their corresponding loss scores. These 12 samples are used to to train the BO and SGD algorithms.}
\label{samples}
\begin{center}
\begin{scriptsize}
\begin{sc}
\begin{tabular}{lcccr}
\toprule
\multicolumn{1}{c}{\textbf{\begin{tabular}[c]{@{}c@{}}Training\\ Sample \#\end{tabular}}} & 
\multicolumn{1}{c}{\textbf{\begin{tabular}[c]{@{}c@{}}Pressure\\ {[}MPa{]}\end{tabular}}} & 
\multicolumn{1}{c}{\textbf{\begin{tabular}[c]{@{}c@{}}Frequency\\ {[}Hz{]}\end{tabular}}} & 
\multicolumn{1}{c}{\textbf{\begin{tabular}[c]{@{}c@{}}Speed\\ {[}mm/s{]}\end{tabular}}} & 
\textbf{Score}  \\ \midrule

1 & 0.090 & 20.4 & 336 & 0.462\\
2 & 0.038 & 16.7 & 393 & 0.422\\
3 & 0.048 & 24.4 & 432 & 0.356\\
4 & 0.118 & 18.0 & 483 & 0.566\\
5 & 0.064 & 36.1 & 510 & 0.309\\
6 & 0.140 & 22.0 & 561 & 0.679\\
7 & 0.079 & 26.9 & 642 & 0.370\\
8 & 0.054 & 30.9 & 675 & 0.358\\
9 & 0.021 & 39.8 & 705 & 0.358\\
10 & 0.129 & 31.9 & 774 & 0.570\\
11 & 0.101 & 34.2 & 822 & 0.501\\ 
12 & 0.113 & 29.4 & 861 & 0.361\\
\bottomrule
\end{tabular}
\end{sc}
\end{scriptsize}
\end{center}
\vspace{-20pt}
\end{table}

\subsection{Bayesian Optimization} 

BO utilizes a surrogate model and acquisition function to efficiently sample a complex $N$-dimensional parameter space by updating prior hypothesis with incoming new data to converge on an optimum \cite{Shahriari2016}. Through efficient sampling and hyperparameter tuning, BO elicits rapid convergence to an optimum using few samples, thus, promoting high-throughput experimentation and reducing waste material \cite{rasmussen2003gaussian, bergstra2012, snoek2012}. In this work, we build our BO model from the tuned settings of Sun \etal \cite{Sun2020} by using a Gaussian process surrogate model, expected improvement (EI) acquisition function, Matern 5/2 kernel with automatic relevance detection, and jitter value of 0.01. The EI acquisition function is used to balance model exploration and exploitation \cite{Zhang2020}. Other acquisition functions were considered for this study such as pure exploration (PE) which considers only exploration in its decision-making policy and maximum variance (MV) which uses exploration in tandem with uncertainty measures and the surrogate model in its decision-making policy. However, the PE and MV acquisition functions require a higher volume of samples to properly and quickly converge due to their tendency to explore a parameter space. EI more informatively guides BO to search regions of our 3-dimensional parameter space that both generate high reward for finding low loss scores and have high uncertainties \cite{Brochu2010,Gongora2020}. Therefore, to achieve an optimum using the fewest samples, the EI acquisition function is used in this study:
\begin{equation} 
\label{EI}
\textbf{x}=\underset{\textbf{x}}{\operatorname{argmin}}\mathbb{E}(\max(0,f^*-f(\textbf{x})) ,
\end{equation}
where $\textbf{x}$ is the improvement of $f^*-f(\textbf{x})$ and $f^*$ is the current minimum value of the evaluated function $f(\textbf{x})$.

BO searches the parameter space where the acquisition value is high, meaning that both the prediction uncertainty and the posterior mean of the objective are high (high objective corresponds to a low loss score in our case since the objective function is quantified as a loss, \textit{i.e.}, the loss score) \cite{Brochu2010,Shahriari2016}. To begin the parameter space search, posterior probabilities are computed for each of the 12 initialized sample conditions using their loss scores, shown in Table \ref{samples}. An optimum printing condition is suggested where the acquisition value, $\textbf{x}$, is maximized, thus, minimizing the loss score from Equation \ref{cv-loss}. The new suggested data are iteratively synthesized in parallel with this acquisition function search until convergence on an optimum within the manifold is achieved. We compare the accuracy and speed of finding this optimum using BO and $k$-fold cross-validated SGD. Achieving accurate and rapid convergence of thin film morphologies via data fusion demonstrates the potential for high-throughput development of complex materials and systems. Reducing the number of synthesized samples required to explore a parameter space saves time and resources.

\subsection{Stochastic Gradient Descent}
To baseline BO in loop for both optimization accuracy and speed, we use a $k=10$ fold cross-validated SGD in loop model. The importance of comparing the performance of these two algorithms is to understand which tool provides the best performance of high-throughput data fusion experimentation while minimizing the number of synthesized samples required to achieve optimum convergence.

In this paper, we use $k=10$ fold SGD to minimize the mean ridge regression error between the actual and model predicted loss scores over $k-1$ folds of training data and the $k^{\textrm{th}}$ fold of validation data, $D=D_{\textrm{train}} \cup D_{\textrm{val}}$ \cite{Kiwiel2001}:
\begin{equation}
\label{w_star}
	\begin{aligned}
	\Theta^*&=\underset{\Theta}{\operatorname{argmin}} \frac{1}{\lvert D\rvert} \sum_{(X,y) \in D}
	\left(L^{(i)}-L_{\textrm{pred}}^{(i)} \right)^2	+\lambda^* \norm{ \Theta}^2,
	\end{aligned}
\end{equation}
where $\Theta^*$ is the best hypothesis classifier after cross validation,
$L$ is the actual loss score, $L_{\textrm{pred}}$ is the predicted loss score, $\lambda^*$ is the best regularization parameter found from cross validation, and $\Theta$ is the weight matrix for each associated input set of hardware conditions, $X$. Predicted loss scores, $L_{\textrm{pred}}$, are computed using the following:
\begin{equation} 
\label{y_pred}
L_{\textrm{pred}}=\Theta^T \cdot X + \Theta_0 ,
\end{equation}
where $\Theta_0$ is the bias. To run 10-fold cross-validation SGD, the training set of 12 droplet images are partitioned into six sub-image and then each of these sub images are augmented by one of the seven transformations before passing through the model: (1) no transformation, (2) rotate 90\textdegree, (3) rotate 180\textdegree, (4) rotate 270\textdegree, (5) mirror vertically, (6) mirror horizontally, or (7) transpose.

From the best SGD hypothesis classifier, $\Theta^*$, a new synthesis condition is obtained that minimizes the loss score. Similar to the BO loop, a new sample is printed using these synthesis conditions and then the algorithm is retrained. This process is repeated until the model converges on an optimum synthesis condition. The converged optimum from the SGD model is compared to that of the BO model to assess the predictive capabilities of each algorithm. Additionally, each step of this data fusion process is timed such that the convergence speed of both algorithms is compared to elicit high-throughput experimentation.

\section{Results}

The convergence time, prediction accuracy, and required sample volume of BO and SGD optimum convergence are shown in Figure \ref{fig:layer_convergence} and Table \ref{comp_time}. BO converges on an optimum in 4 iterations, each taking 162.1 seconds, for a total of 648.6 seconds with a prediction accuracy of $\textrm{RMSE} = 7.36\mathrm{e}{-4}$. 10-fold cross-validated SGD converges on an optimum in 3 iterations, each taking 430.6 seconds, for a total of 1291.8 seconds with a prediction accuracy of $\textrm{RMSE} = 0.242$. Hence, BO requires 16 total samples for optimum convergence and is nearly 2x faster than SGD.

\subsection{Trained Model Search Space} \label{search_space}

\begin{figure}
\centering
\includegraphics[width=\columnwidth]{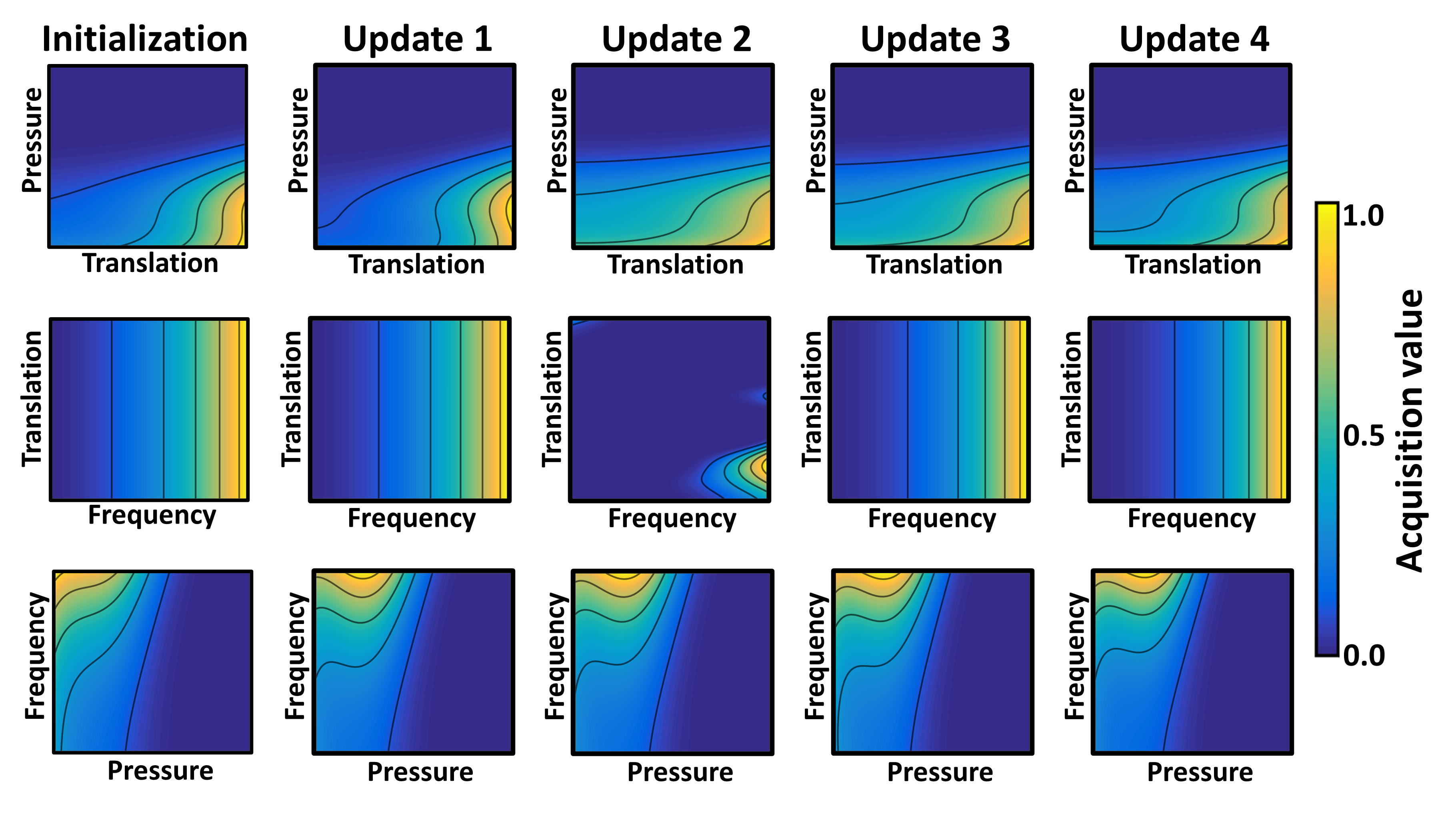}
\caption{\textit{Expected improvement} acquisition function values for each update of the BO model with new data until convergence by update 4. Each plot is a 2D cross-section of the 3D parameter space such that the acquisition values are shown for the slices: pressure $\times$ frequency, frequency $\times$ translation, and translation $\times$ pressure. Higher acquisition values represent parameter values with higher objective posterior means and higher uncertainties. The BO model is more likely to suggest optima from these regions. Each update slightly changes the acquisition value of the parameter space because new sample data is introduced. After introducing the optimum suggested from update 1, the acquisition function narrows its region of expected improvement in the frequency $\times$ translation cross-section in update 2 until the suggested optimum from update 2 expands the region back in update 3.}
\label{fig:BO_acq}
\vspace{-10pt}
\end{figure}

Iterative updates to the BO model search spaces are shown by the acquisition values of EI in Figure \ref{fig:BO_acq}. Higher acquisition values indicate locations within the parameter space that are expected to yield a higher improvement for finding a minimum loss score \cite{Brochu2010,Gongora2020}. 

Figure \ref{fig:BO_acq} illustrates that the search region of BO during initialization is concentrated in the low pressure values, high frequency values, and high translation speed values. Therefore, from this trained model, we expect BO to iteratively sample optima condition values within those regions of low pressure, high frequency, and high speed after each retraining until optimum convergence is achieved; as new optima are added to the training set and BO is retrained, slight variation in the acquisition is noted throughout the updates. We demonstrate that this sampling pattern is generally true throughout the experiment, as shown by the acquisition values plotted for updates 1--4 in Figure \ref{fig:BO_acq}.

SGD defines its search based on regression weights for each parameter in the space, that minimize the mean square error between predicted and actual loss scores over the $k-1$ folds of training data and $k^{\textrm{th}}$ fold of validation data to achieve $\epsilon\leq0.01$. In 160 iterations, SGD converges to a hypothesis classifier with weights $\Theta=(0.121,0.313,0.148)$. This hypothesis classifier is used to determine the synthesis conditions which generate the lowest estimated loss score and the classifier is iteratively updated as new optima are added to the training set.

\subsection{Optimum Convergence}

\begin{figure}

\centering
\subfigure[Converged BO 1D Manifold]{
\includegraphics[width=0.4\textwidth]{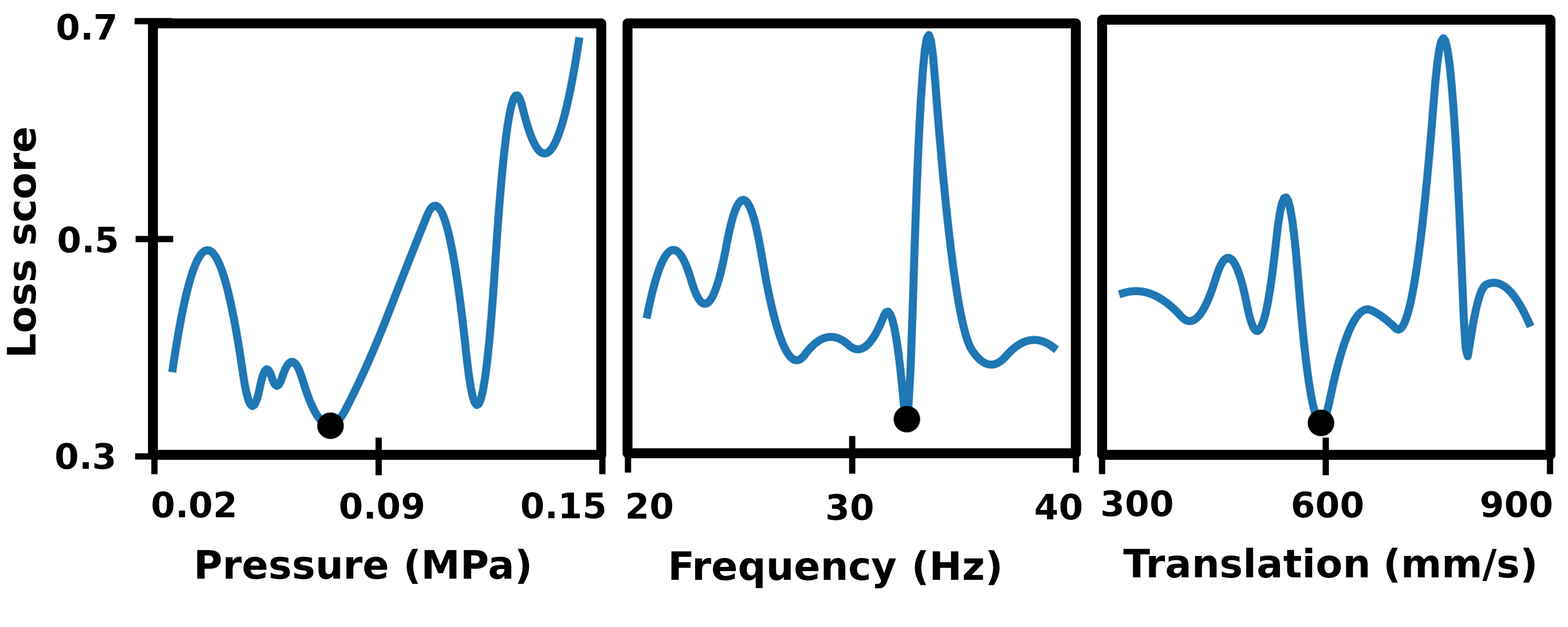}
\label{fig:BO_1D}
}

\subfigure[Converged SGD 1D Manifold]{
\includegraphics[width=0.4\textwidth]{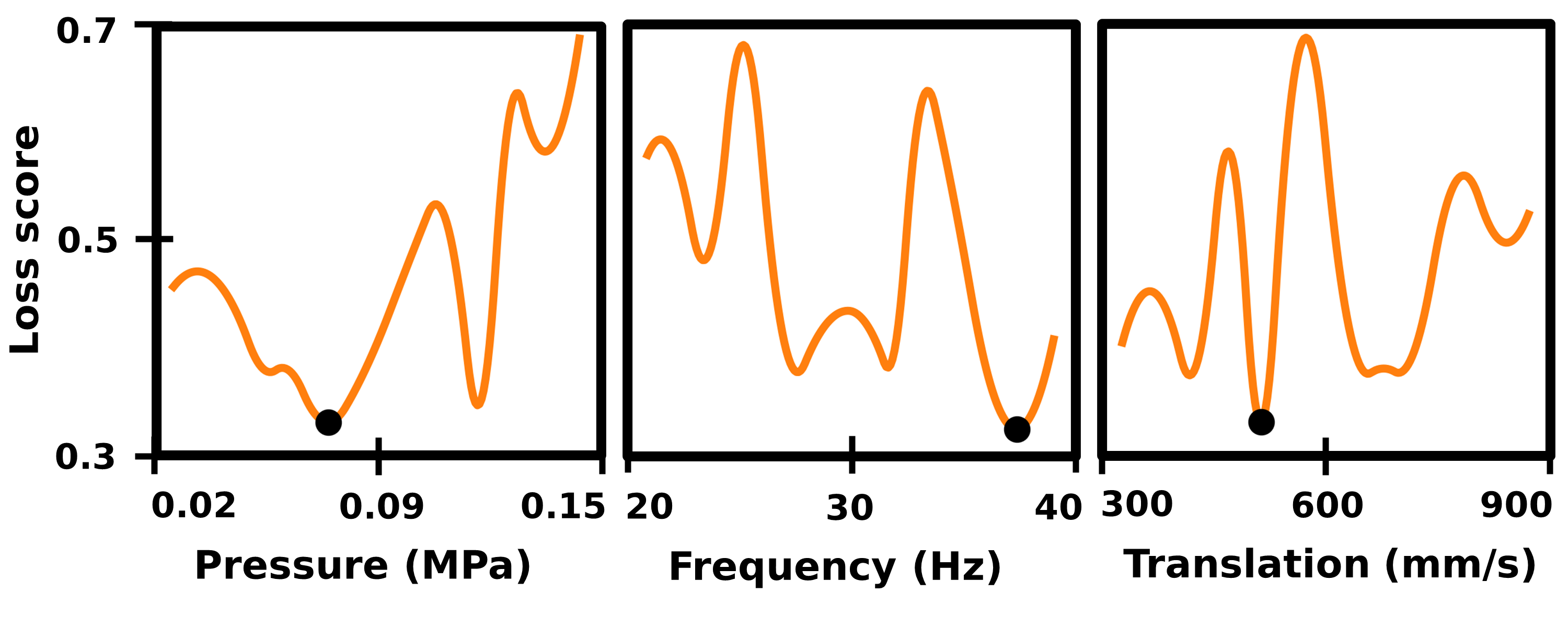}
\label{fig:SGD_1D}
}

\caption{1D interpolated manifold representation of each synthesis parameter and its associated loss score after (a) BO convergence in 4 updates and (b) SGD convergence after 3 updates. Each plot represents a change in computed loss score given the training synthesis conditions and suggested synthesis conditions by BO and SGD. The BO manifold contains 4 suggested optima and the SGD manifold contains 3 suggested optima. The black point indicates the synthesis condition with the lowest loss score.}
\label{fig:1D_manifold}
\vspace{-10pt}
\end{figure}

\begin{figure*}
\centering
\subfigure[BO of Inkjet Droplets]{
    \includegraphics[width=.75\textwidth]{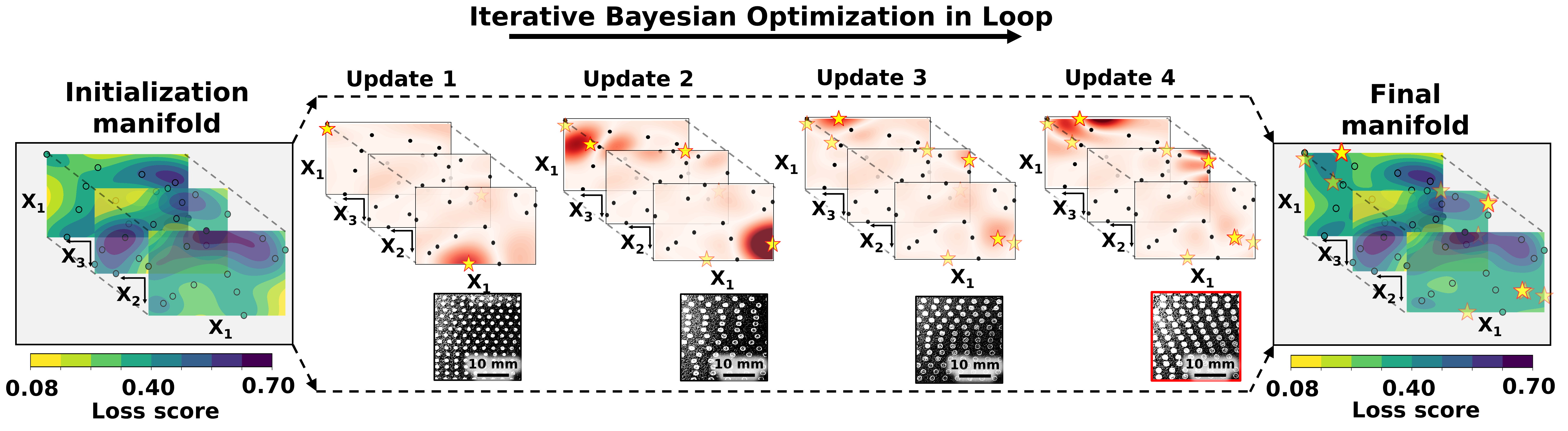}
    \label{fig:BO_layer}
}

\subfigure[SGD Optimization of Inkjet Droplets]{
    \includegraphics[width=.75\textwidth]{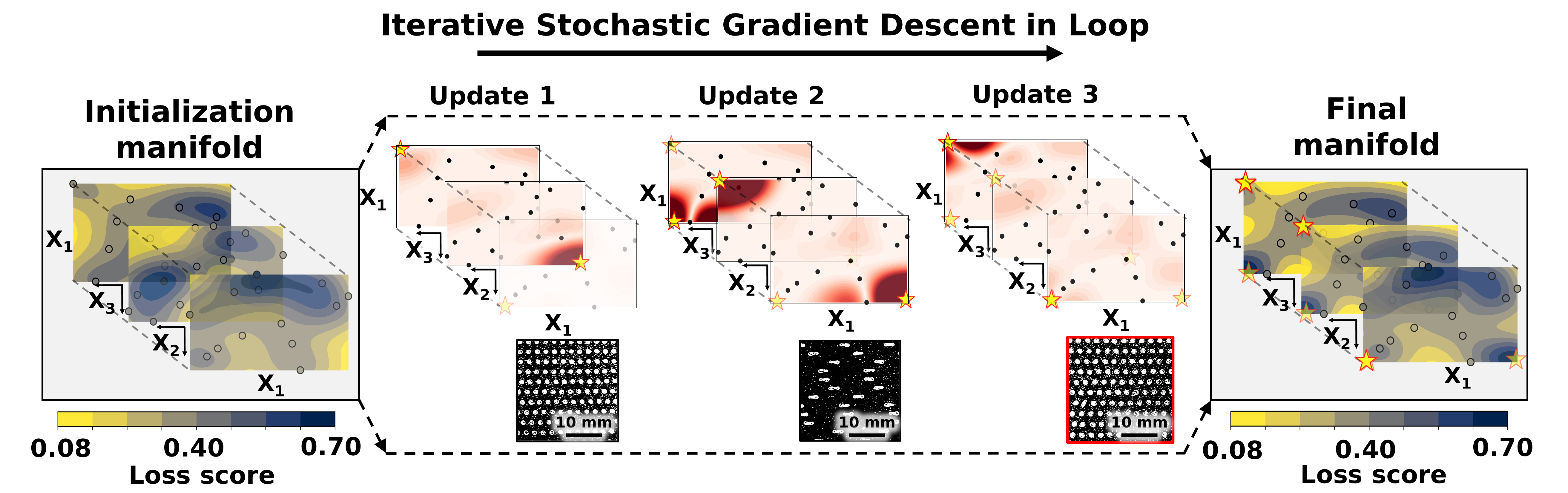}
    \label{fig:SGD_layer}
}

\caption{The process flow of (a) BO in loop and (b) SGD in loop achieving convergence on an optimum droplet structure and synthesis condition. The manifolds are illustrated as 2-dimensional cross-sections of the 3-dimensional parameter space: $x_1 \times x_2 \times x_3$ corresponding to pressure, frequency, and translation speed, respectively. The experimentally synthesized inkjet droplet structure for each model update is shown below the manifolds. The circle scatter points represent the 12 LHS-initialized conditions and stars represent the predicted optima. The initialized manifolds show the Gaussian interpolated loss score for the 12 initial conditions. From the initialization manifold, BO and SGD predict an optimum synthesis condition which is then used to generate a droplet sample, shown under update 1. The loss score delta manifold is illustrated as the red heatmap and quantifies the difference in loss score across the manifold from one update to the next, darker colors represent a larger difference in loss score between updates. Once the algorithm suggests the same synthesis condition for several updates in a row, the loss score delta manifold approaches zero change. BO converges to an optimum droplet structure in 4 updates and SGD in 3 updates -- the optimum droplet structure is highlighted in red. The final manifold illustrates the Gaussian interpolated loss scores of the aggregated training data and the predicted optima.}
\label{fig:layer_convergence}
\vspace{-10pt}
\end{figure*}

Convergence on an optimum droplet structure and synthesis condition of the BO and SGD loops is shown in Figure \ref{fig:layer_convergence}. BO converges on an optimum droplet structure in 4 updates and SGD converges on an optimum droplet structure in 3 updates. The updated relationships between synthesis variables and loss score are shown in Figure \ref{fig:1D_manifold}.

In Figure \ref{fig:layer_convergence}, the loss score delta manifolds are shown for each update alongside the images of the fabricated optimum droplet structure for that update. For each consecutive update, the difference in loss score from the previous step is computed across the parameter space manifold. BO begins to suggest the same optima as the prior step by update 4 and SGD beings to suggest the same optima as the prior step by update 3. At this point, the loss score difference across the manifold approaches zero, thus, convergence on an optimum is achieved. After optimum convergence is achieved, the final manifold is shown in Figure \ref{fig:layer_convergence} with the optimum synthesis conditions illustrated by the star.

For the BO in loop process in Figure \ref{fig:BO_layer}, the optimum sample condition suggested by the first update has a value of low pressure, high frequency, and average speed, as expected from the acquisition function in Figure \ref{fig:BO_acq}. After synthesizing and imaging the suggested optimum, it receives an actual loss score of 0.359, which is higher than its predicted value of 0.147. Then, the algorithm is retrained on the 12 LHS-initialized samples plus the suggested optimum of the previous update to suggest an updated optimum in the next update. By the fourth iteration of updating the priors with new data, the algorithm converges to an optimum that is suggested for two updates in a row: $(x_1=0.053  \textrm{MPa}, x_2=39.8 \textrm{Hz}, x_3=260 \textrm{mm/s},L_{\textrm{pred}}=0.342,L=0.341)$. The droplet structure experimentally synthesized using these printing conditions is illustrated by red-outlined droplet image in update 4 of Figure \ref{fig:BO_layer}. BO starts its optimum prediction with an $\textrm{RMSE}=0.150$ in update 1 and then ends with an optimum prediction $\textrm{RMSE} = 7.36\mathrm{e}{-4}$ after it converges by update 4.

For the SGD in loop as shown in Figure \ref{fig:SGD_layer}, the optimum sample condition suggested by the first update has a value of low pressure, high frequency, and low speed, which is similar to the optimum suggested by the BO except the BO suggests to search within high speed values instead of low. After synthesizing and imaging this initial optimum, it receives an actual loss score of 0.354, which is significantly higher than the predicted value of 0.003 by SGD. After being retrained on the previous optimum, SGD suggests a new optimum at low pressure, low frequency, and high speed, which generates a non-uniform and non-connected thin film with a loss score of 0.694, shown in update 2 of Figure \ref{fig:SGD_layer}. On the next update, SGD converges back to previously suggested optimum in update 1: $(x_1=0.021  \textrm{MPa}, x_2=39.8 \textrm{Hz}, x_3=113 \textrm{mm/s},L_{\textrm{pred}}=0.002,L=0.344)$. The droplet structure experimentally synthesized using these printing conditions is illustrated by red-outlined droplet image in update 3 of Figure \ref{fig:SGD_layer}. SGD starts its optimum prediction with an $\textrm{RMSE}=0.248$ in update 1 and then ends with an optimum prediction $\textrm{RMSE} = 0.242$ after it converges by update 3.

\subsection{Computation Time}

To elicit high-throughput experimentation, algorithmic computational speed and synthesis speed are important. The computing and synthesis times required to achieve optimum convergence are shown in Table \ref{comp_time} for BO and $k=10$ fold cross-validated SGD. Overall, the BO in loop method is 2x faster than the SGD in loop method to attain convergence on an optimum synthesis condition and droplet structure.

\begin{table}[t]
\small
\caption{Average time spent on each process step per update of BO in loop and SGD in loop data fusion methods, measured in seconds. $\textrm{k}=10$ SGD is trained and validated on 504 sub-images augmented from the 12 full images -- this increases the read image and training time but decreases the average loss score computation time because the total pixel count of each sub-image is smaller than that of the full image. BO takes 4 updates to converge, $\implies 162.1\textrm{s}\times4=648.4\textrm{s}$ and SGD takes 3 updates to converge, $\implies 430.6\textrm{s}\times3=1291.8\textrm{s}$.}
\label{comp_time}
\begin{center}
\begin{footnotesize}
\begin{sc}
\begin{tabular}{lcccr}
\toprule
\textbf{Process Step} 
& \multicolumn{1}{c}{\textbf{\begin{tabular}[c]{@{}c@{}}BO Time\\ {[}s{]}\end{tabular}}} 
& \multicolumn{1}{c}{\textbf{\begin{tabular}[c]{@{}c@{}}k=10 SGD Time\\ {[}s{]}\end{tabular}}}
\\ \hline
\multicolumn{1}{l|}{Read Images} & 0.1 & 3.1 \\
\multicolumn{1}{l|}{Compute Score} & 26.6 & 15.5 \\
\multicolumn{1}{l|}{Train Model} & 0.4 & 277.0 \\
\multicolumn{1}{l|}{Printer Set Up} & 70.0 & 70.0 \\
\multicolumn{1}{l|}{Print Droplets} & 30.0 & 30.0 \\
\multicolumn{1}{l|}{Image Droplets} & 35.0 & 35.0 \\ \hline
Total per Update & 162.1 & 430.6\\
\textbf{Total for Convergence}& \textbf{648.4}  & \textbf{1291.8} \\
\bottomrule
\end{tabular}
\end{sc}
\end{footnotesize}
\end{center}
\vspace{-20pt}
\end{table}

Table \ref{comp_time} shows the computing times for each step in SGD and BO per update as well as the time taken to synthesize and image the predicted optimum samples. The total optimum convergence runtime of BO is 2x faster than the total optimum convergence runtime of SGD. BO takes 4 updates (162.1 seconds per update) to achieve convergence and SGD takes 3 updates (430.6 seconds per update) to achieve convergence. Comparing only the model training times, BO is 693x faster than SGD with $k=10$, where a significant amount of SGD training time is spent doing cross validation. For comparison purposes, if we assume we know the optimum $\lambda^*$ value \textit{a priori}, SGD can be reduced to $k=1$ in which the BO training time is still 157x faster than the SGD training time. Hence, illustrating the importance of reducing the number of samples for data fusion experimentation to achieve optimum convergence.

\subsection{Discussion}

In this paper we propose a method for online optimization of inkjet printer parameters such that we precondition that hardware for semiconductor synthesis and characterization without the intervention of a domain expert. Using this proposed method, we demonstrate accurate convergence on hardware printer conditions in 650 seconds that optimize droplet geometric uniformity and yield using 16 or fewer total samples. Autonomously optimizing inkjet hardware conditions using computer vision-driven machine learning in loop democratizes the ability to synthesize stable, high-performance semiconductors.

Both BO and SGD models are initialized using 12 droplet samples experimentally synthesized by a custom inkjet printer. The printing conditions used to synthesize these samples are obtained by LHS of the inkjet parameter space (pressure $\times$ frequency $\times$ speed) to preserve variability of the parameter space. After this initialization, the BO and SGD models search for synthesis conditions within this parameter space that optimize the droplet structures according to our defined objective function: maximize droplet uniformity and maximize yield. The EI acquisition function of BO searches for optimized droplet structures within low pressure, high frequency, and high translation speed fabrication conditions. The 10-fold cross-validated hypothesis classifier of SGD guides a search for optimized droplet structures within low pressure, high frequency, and low translation speed fabrication conditions.

To achieve optimum convergence, BO in loop takes 4 updates of optimization whereas SGD in loop takes 3 updates. BO converges on the optimum conditions of $(x_1=0.053  \textrm{MPa}, x_2=39.8 \textrm{Hz}, x_3=260 \textrm{mm/s})$ and SGD converges on the optimum conditions of  $(x_1=0.021  \textrm{MPa}, x_2=39.8 \textrm{Hz}, x_3=113 \textrm{mm/s})$. To achieve this optimum convergence, BO in loop is 2x faster than SGD in loop, where the total BO convergence time is $648.4\textrm{s}$ and the total SGD convergence time is $1291.8\textrm{s}$. Once convergence is attained, BO has an optimum prediction accuracy of $\textrm{RMSE} = 7.36\mathrm{e}{-4}$ and SGD has an optimum prediction accuracy of $\textrm{RMSE}=0.242$. Although SGD in loop converges in fewer data fusion iterations than BO in loop, BO predicts optima more accurately and is 693x faster to retrain than SGD. However, without the use of a computer vision-driven machine learning in loop method or the intervention of a domain expert, experimental synthesis of this parameter space could require hundreds of samples and days of manual labor to fully search and discover an optimum inkjet hardware condition. We demonstrate that with no knowledge \textit{a priori} of a 3D parameter space, BO offers an accurate and fast method to optimize inkjet-deposited droplet structures using only 12 training samples.

In this paper, we demonstrate that using a computer vision-driven Bayesian optimization process in loop with experimental synthesis speeds-up optimization time using very few samples. Through this accelerated and economical optimization of inkjet hardware, we precondition an experimental setup autonomously and in real-time without the intervention of a domain expert. Thus, enabling researchers to conduct high-throughput experimentation of semiconductors to discover compositions of optimal properties (\textit{e.g.} high efficiency, low degradation, and high electron mobility) -- which is a gap in current semiconductor optoelectronic engineering. 

\section{Conclusions}

We present a method for accurate, fast, and economical experimental hardware optimization using computer vision-driven machine learning. Using this automated method of hardware process optimization, researchers can easily precondition experimental setups which would otherwise require iterative tuning by a domain expert. In this paper, we compare the accuracy and speed of two machine learning loops: (1) Bayesian optimization and (2) stochastic gradient descent of convergence on geometrically uniform and high-yield inkjet deposited droplet structures. A two-level model is used to detect and derive a loss score for these droplet structures such that the machine learning method discovers the inkjet hardware conditions that minimize the loss score of the structures. Using 12 initial training samples, both methods converge on a similar optimum droplet structure -- obtained through parallel iterative experimental synthesis and model retraining. However, Bayesian optimization is demonstrated to converge 2x faster and with a higher accuracy than stochastic gradient descent. The significance of demonstrating the accuracy and speed of this economical hardware process optimization tool is in its utility for researchers to precondition experimental setups without a domain expert. By optimizing droplet structures to have high uniformity and yield, this method elicits high-throughput synthesis and characterization of semiconductor compositions such that time and material resources are reduced to find stable, high-performance semiconductor compositions.

{\small
\bibliographystyle{plain}
\bibliography{bibliography}
}

\end{document}